%% file: 0_Main.tex
\documentclass{article}
\usepackage[utf8]{inputenc}
\usepackage[numbers]{natbib}
\usepackage{subcaption}
\DeclareUnicodeCharacter{202F}{\,}
\usepackage{makecell}
\usepackage{amsmath}
\usepackage{graphicx}
\usepackage{epstopdf}
\usepackage{multirow}
\usepackage{times}
\usepackage{latexsym}
\usepackage{amssymb}
\usepackage{newunicodechar}
\newunicodechar{，}{,}
\usepackage[T1]{fontenc}
\usepackage[utf8]{inputenc}
\usepackage{authblk}
\usepackage{bm}
\usepackage{booktabs}
\usepackage[table]{xcolor}  % required for \rowcolor{gray!15}
\usepackage{colortbl}
\usepackage{microtype}
\usepackage{float}
\usepackage[lined,boxed,commentsnumbered]{algorithm2e}
\usepackage{inconsolata}
\usepackage[]{changes}
\usepackage{hyperref}
\usepackage{url}
\input{math_commands.tex}

\title{OverdoseMoE: A Multi-Expert Framework for Opioid Overdose Risk Prediction}
 \author[1,3]{Mingchen Li}
  \author[1,3]{Rohan Pandey}
  \author[2,3]{Junhui Qian}
  \author[2,3]{Feiyun Ouyang}
  \author[1,3]{Sunjae Kwon}
  \author[1,2,3]{Hong Yu}

\affil[1]{Manning College of Information and Computer
Sciences, UMass Amherst, Amherst, MA, USA}
\affil[2]{Miner School of Computer and Information
Sciences, UMass Lowell, Lowell, MA, USA}
\affil[3]{Center for Healthcare Organization and Implementation Research, VA Bedford Health Care, Bedford, MA, USA}
\date{}
\begin{document}

\maketitle
\begin{center}
% \small $^*$These authors contributed equally.
\end{center}
\input{1_abs}

\input{2_introduction}

\input{3_Results}
\input{4_Discussion}
\input{5_method}

\input{6_Other}

% , go at the end of the paper.

\clearpage  % flush all deferred floats before bibliography
\bibliography{biblatex-nature}
\bibliographystyle{naturemag}
% \bibliographystyle{naturemag}
% \bibliography{references}  % replace with your actual .bib filename (e.g., myrefs.bib)
% \bibliography{science_template}

\clearpage  % flush any remaining floats before supplementary
\appendix
\input{8_appendix}

\end{document}

%% file: math_commands.tex
\usepackage{amsmath,amsfonts,bm}

\def\eqref#1{equation~\ref{#1}}
\def\1{\bm{1}}

\DeclareMathAlphabet{\mathsfit}{\encodingdefault}{\sfdefault}{m}{sl}
\SetMathAlphabet{\mathsfit}{bold}{\encodingdefault}{\sfdefault}{bx}{n}

%% file: 1_abs.tex
\begin{abstract}

Opioid overdose remains a major clinical and public health burden, motivating improved approaches for identifying patients at high risk. In this study, we investigate opioid overdose risk prediction from longitudinal clinical histories in 78,660 Veterans Health Administration (VHA) patients with opioid use disorder diagnosed between 2017 and 2019. To this end, we develop \textsc{OODMamba} and \textsc{OODQwen} through continued pretraining on longitudinal ICD-code sequences followed by task-specific fine-tuning. Building on the stronger Qwen-based predictors, we further propose \textsc{OverdoseMoE}, a multi-expert framework that integrates models of different scales using complementary expert-weighting strategies.
Experimental results show that our adapted models consistently outperformed the evaluated baselines, with \textsc{OODQwen} increasing AUPRC by 14.7\%. \textsc{OverdoseMoE} further extended this gain to 17.9\% over the strongest baseline.
Among patients ranked in the top 5\% of predicted risk, \textsc{OverdoseMoE} identified a substantially enriched high-risk subgroup, achieving a PPV of 25.38\% while retaining meaningful recall, suggesting potential utility for targeted clinical intervention. To assess generalizability beyond the VHA cohort, we further evaluated the model on an independent MIMIC-IV cohort, where complementary weighting strategies showed advantages across different performance measures. Together, these results show that combining longitudinal ICD-code modeling with multi-expert integration improves opioid overdose risk stratification and supports robust generalization across independent patient cohorts\footnote{Code associated with this study is available at
\url{https://github.com/ToneLi/OverdoseMoE/}.}.

\end{abstract}

%% file: 2_introduction.tex
\section{Introduction}
Opioid overdose remains a major public health challenge in the United States.
Although overdose mortality has declined in recent years, the latest CDC provisional estimates indicate approximately 67,798 drug overdose deaths in the United States during the 12 months ending in March 2026, representing a 12.3\% decline from the previous year.
\footnote{\url{https://www.cdc.gov/overdose-prevention/data-research/facts-stats/index.html}}.
Importantly, opioid overdose is preventable. Overdose education and naloxone distribution
have been associated with reductions in opioid-related mortality~\cite{walley2013opioid}, while medications for
opioid use disorder, particularly buprenorphine and methadone, are associated with lower
mortality following nonfatal overdose~\cite{larochelle2018medication}. Clinical guidelines further recommend targeted
risk-mitigation strategies, including naloxone provision, prescription drug monitoring,
appropriate toxicology screening, and evidence-based treatment for opioid use disorder,
particularly among patients at increased risk~\cite{dowell2022cdc}. These effective interventions underscore
a critical challenge for health systems: identifying patients at elevated near-term overdose
risk early enough to prioritize preventive care.

Prior studies have demonstrated the feasibility of predicting opioid overdose from routinely collected EHR and administrative data. Lo-Ciganic et al. applied logistic regression, random forest, gradient boosting, and deep neural networks to predict subsequent opioid overdose among Medicare beneficiaries with opioid prescriptions~\cite{lociganic2019evaluation}, while Dong et al. evaluated traditional machine-learning and neural models using longitudinal claims and EHR histories~\cite{dong2019machine}. More recent work has incorporated temporal modeling: Dong et al. developed LSTM-based models to capture longitudinal clinical patterns for opioid overdose risk prediction~\cite{dong2021predicting}, and Rahman et al. compared classical and machine-learning survival models, including random survival forests and survival SVMs, for imminent overdose prediction~\cite{rahman2026imminent}. Collectively, these studies demonstrate the value of longitudinal EHR data for overdose risk prediction. However, most existing approaches rely on structured or engineered clinical features rather than large-scale continued pretraining over longitudinal diagnosis sequences.

Recent advances in pretrained EHR and clinical language models provide new opportunities to learn representations directly from longitudinal patient histories. TransformEHR demonstrated that large-scale pretraining over longitudinal EHR sequences can improve downstream clinical outcome prediction~\cite{yang2023transformehr}, while ClinicalMamba showed that state-space architectures can efficiently model long clinical sequences at scale~\cite{yang2024clinicalmamba}. However, these models were developed primarily for general clinical representation learning rather than being specifically adapted to longitudinal diagnosis trajectories associated with opioid overdose. The use of continued pretraining on large-scale diagnosis histories followed by task-specific fine-tuning for opioid overdose prediction therefore remains underexplored.

Ensemble and multi-model approaches have also been investigated for opioid overdose prediction. For example, Ripperger et al~\cite{ripperger2022ensemble}. trained multiple random-regression-forest weak learners using aggregated features derived from prescription drug monitoring data, hospital discharge records, diagnoses, medications, healthcare utilization, and socioeconomic indicators, and subsequently combined their predictions using several ensemble strategies. More recently, HIBERT~\cite{ding2024hibert} employed multiple specialized BERT modules for different EHR feature categories, including diagnoses, laboratory tests, and medications, pretrained these modules using unlabeled EHR data, and integrated their representations for opioid overdose prediction. Nevertheless, it remains underexplored whether general-purpose pretrained models can be further adapted through continued pretraining on large-scale longitudinal diagnosis text, fine-tuned for opioid overdose prediction, and subsequently integrated across model scales within a multi-expert framework.

To address these gaps, we develop \textsc{OODMamba} and \textsc{OODQwen}, two diagnosis-adapted models for 180-day opioid overdose risk prediction from patients' preceding one-year longitudinal diagnosis histories. Both models are obtained through continued pretraining on large-scale longitudinal ICD-coded diagnosis sequences followed by task-specific fine-tuning. Building on the stronger Qwen-based predictors, we further propose \textsc{OverdoseMoE}, a multi-expert framework that integrates models of different scales using three expert-weighting strategies. We evaluate our approach on a nationwide Veterans Health Administration cohort against classical machine-learning methods, deep sequential models, and pretrained language models, with additional analyses of high-risk patient identification, inference efficiency, and cross-cohort generalizability.\textsc{OverdoseMoE} achieved an AUPRC of 25.17 and an AUROC of 69.49, while attaining a PPV of 25.38\% among patients ranked in the top 5\% of predicted risk. Our \textsc{OODMamba} additionally demonstrated strong cross-cohort generalizability on MIMIC-IV.

Our contributions are fourfold. First, we construct a large-scale longitudinal diagnosis dataset from a nationwide VHA cohort for 180-day opioid overdose risk prediction, representing patients' preceding clinical histories as chronologically ordered ICD-based diagnosis text. Second, we develop two diagnosis-adapted models, \textsc{OODMamba} and \textsc{OODQwen}, through continued pretraining on large-scale longitudinal diagnosis histories followed by task-specific fine-tuning. Third, our experiments demonstrate that diagnosis-specific continued pretraining consistently improves downstream overdose prediction over directly fine-tuned counterparts, supporting the value of adapting pretrained models to longitudinal diagnostic trajectories. Finally, we propose \textsc{OverdoseMoE}, a multi-expert framework that integrates Qwen-based predictors of different scales, and show that multi-expert learning further improves predictive performance and high-risk patient identification beyond individual models.

%% file: 3_Results.tex
\section{Results}

\subsection{Cohort characteristics}

The study included 62,928 patients in the VHA training set, 7,866 in the
validation set, and 7,866 in the held-out test set, together with 3,971
patients in the external MIMIC-IV cohort
(Table~\ref{tab:dataset_stats}). The prevalence of 180-day opioid overdose
was similar across the VHA splits, ranging from 5.59\% to 6.01\%, whereas
the prevalence in MIMIC-IV was substantially lower at 1.38\%.
Demographic characteristics were also comparable across the VHA training,
validation, and test sets but differed markedly between VHA and MIMIC-IV
(Table~\ref{tab:demographics}). The MIMIC-IV cohort was younger overall and
had a substantially higher proportion of women (46.78\%) than the VHA
cohorts (approximately 8\%). These differences highlight the demographic
shift between the development and external cohorts and provide a distinct
setting for evaluating cross-cohort transportability.
\begin{table}[ht]
\centering
\small
\renewcommand{\arraystretch}{1.15}

\begin{tabular*}{0.82\linewidth}{@{\extracolsep{\fill}}llrr@{}}
\toprule
\textbf{Data source} &
\textbf{Split} &
\textbf{N} &
\textbf{Overdose, n (\%)} \\
\midrule
VHA      & Training   & 62{,}928 & 3{,}724 (5.92\%) \\
VHA      & Validation & 7{,}866  & 440 (5.59\%) \\
VHA      & Test       & 7{,}866  & 473 (6.01\%) \\
MIMIC-IV & External   & 3{,}971  & 55 (1.38\%) \\
\bottomrule
\end{tabular*}

\caption{
Cohort size and 180-day opioid-overdose prevalence in the VHA development
cohort and the external MIMIC-IV cohort. The VHA cohort was partitioned
at the patient level into training, validation, and held-out test sets.
}
\label{tab:dataset_stats}
\end{table}

% \begin{table}[ht]
% \centering
% \scriptsize
% \renewcommand{\arraystretch}{1.12}
% \setlength{\tabcolsep}{8pt}
% \begin{tabular}{llrr}
% \toprule
% \textbf{Data source} & \textbf{Split} &
% \textbf{N} & \textbf{Overdose, n (\%)} \\
% \midrule
% VHA      & Training   & 62{,}928 & 3{,}724 (5.92\%) \\
% VHA      & Validation & 7{,}866  & 440 (5.59\%) \\
% VHA      & Test       & 7{,}866  & 473 (6.01\%) \\
% MIMIC-IV & External   & 3{,}971  & 55 (1.38\%) \\
% \bottomrule
% \end{tabular}

% \caption{
% Cohort size and 180-day opioid-overdose prevalence in the VHA development
% cohort and the external MIMIC-IV cohort. The VHA cohort was partitioned
% at the patient level into training, validation, and held-out test sets.
% }
% \label{tab:dataset_stats}
% \end{table}

\begin{table}[ht]
\centering
\scriptsize
\renewcommand{\arraystretch}{1.12}
\setlength{\tabcolsep}{5pt}
\resizebox{0.9\textwidth}{!}{%
\begin{tabular}{lrrrr}
\toprule
\textbf{Characteristic} &
\textbf{Training} &
\textbf{Validation} &
\textbf{Test} &
\textbf{MIMIC-IV} \\
\midrule

\multicolumn{5}{l}{\textbf{Age group, years}} \\
18--30
& 3{,}387 (5.38\%)
& 385 (4.89\%)
& 442 (5.62\%)
& 569 (14.33\%) \\

31--40
& 10{,}201 (16.21\%)
& 1{,}352 (17.19\%)
& 1{,}272 (16.17\%)
& 601 (15.14\%) \\

41--50
& 7{,}323 (11.64\%)
& 845 (10.74\%)
& 917 (11.66\%)
& 775 (19.51\%) \\

51--60
& 14{,}327 (22.77\%)
& 1{,}812 (23.03\%)
& 1{,}783 (22.67\%)
& 916 (23.06\%) \\

61--70
& 20{,}332 (32.31\%)
& 2{,}558 (32.52\%)
& 2{,}542 (32.32\%)
& 599 (15.08\%) \\

71--80
& 6{,}169 (9.81\%)
& 770 (9.79\%)
& 782 (9.94\%)
& 313 (7.88\%) \\

80+
& 1{,}189 (1.89\%)
& 144 (1.83\%)
& 128 (1.63\%)
& 198 (4.99\%) \\

\midrule

\multicolumn{5}{l}{\textbf{Sex}} \\
Female
& 4{,}872 (7.74\%)
& 671 (8.53\%)
& 624 (7.93\%)
& 1{,}857 (46.78\%) \\

Male
& 58{,}056 (92.26\%)
& 7{,}195 (91.47\%)
& 7{,}242 (92.07\%)
& 2{,}114 (53.22\%) \\

\midrule

\multicolumn{5}{l}{\textbf{Race/ethnicity}} \\
Not Hispanic or Latino
& 58{,}368 (92.77\%)
& 7{,}312 (92.95\%)
& 7{,}281 (92.56\%)
& 3{,}618 (91.11\%) \\

Hispanic or Latino
& 3{,}490 (5.55\%)
& 425 (5.40\%)
& 473 (6.01\%)
& 198 (4.99\%) \\

Other or unknown
& 1{,}070 (1.70\%)
& 129 (1.64\%)
& 112 (1.42\%)
& 155 (3.90\%) \\

\bottomrule
\end{tabular}
}

\caption{
Demographic characteristics of the VHA training, validation, and held-out
test cohorts and the external MIMIC-IV cohort. Values are reported as
number of patients and percentage within each cohort.
}
\label{tab:demographics}
\end{table}

\subsection{Overall model performance}
\begin{table*}[ht]
\centering
\small
\renewcommand{\arraystretch}{1.12}

\begin{tabular*}{0.90\textwidth}
{@{\extracolsep{\fill}}llccc@{}}
\toprule
\textbf{Family} & \textbf{Model} &
\textbf{AUPRC} & \textbf{AUROC} &
\textbf{PPV} \\
\midrule

\multirow{3}{*}{Classical ML}
& RandomForest        & 10.07 & 64.50 & 6.77 \\
& LogisticRegression  & 11.93 & 59.46 & 3.38 \\
& SVM                 & 7.02  & 54.22 & 8.88 \\

\midrule

\multirow{2}{*}{Deep sequence}
& GRU             & 11.26 & 62.48 & 7.61 \\
& TransformerEHR  & 13.51 & 60.13 & 7.44 \\

\midrule

\multirow{6}{*}{Fine-tuned LM}
& Qwen3-1.7B          & 20.79 & 67.59 & 12.25 \\
& ClinicalMamba-2.8B  & 10.35 & 65.22 & 10.92 \\
& Qwen2.5-3B          & 11.14 & 64.17 & 10.78 \\
& Qwen3-4B            & 21.34 & 66.36 & 12.17 \\
& BioMistral-7B       & 13.22 & 64.09 & 9.80  \\
& Qwen3-8B            & 20.68 & 65.19 & 13.83 \\

\midrule

\multirow{2}{*}{Diagnosis-adapted}
& \textsc{OODMamba-2.8B} & 11.15 & 66.11 & 11.74 \\
& \textsc{OODQwen3-1.7B} & 24.47 & 68.56 & 12.34 \\

\midrule

\multirow{3}{*}{\textsc{OverdoseMoE}}
& \textbf{EW}  & 25.01 & 69.28 & 14.60 \\
& \textbf{PGA} & \textbf{25.17} & \textbf{69.49} &
\textbf{14.95} \\
& \textbf{GLA} & 24.97 & 69.14 & 14.38 \\

\bottomrule
\end{tabular*}

\caption{
Primary predictive performance for 180-day opioid overdose prediction on the
held-out VHA test set. Models are grouped into classical machine-learning
baselines, deep-sequence models, fine-tuned language models,
diagnosis-adapted models, and the proposed \textsc{OverdoseMoE} framework.
\textsc{OverdoseMoE} includes three expert-fusion variants:
\textbf{EW} (Equal-Weight Fusion),
\textbf{PGA} (Prior-Guided Adaptive Fusion), and
\textbf{GLA} (Global-Local Adaptive Fusion).
Risk-stratified PPV and recall at prespecified top-$k$ thresholds are
reported separately in Table~\ref{tab:risk_stratified}.
}
\label{tab:primary_results}
\end{table*}

To evaluate overall predictive performance, we compared the proposed
diagnosis-adapted models and \textsc{OverdoseMoE} with classical
machine-learning models, deep-sequence models, and directly fine-tuned
language models for 180-day opioid overdose prediction
(Table~\ref{tab:primary_results}). Among the individual models,
\textsc{OODQwen3-1.7B} achieved the highest AUPRC of 24.47 and AUROC of
68.56, while Qwen3-8B achieved the highest PPV of 13.83 among the directly
fine-tuned language models. All three \textsc{OverdoseMoE} variants
achieved higher AUPRC and AUROC than the individual models. The
Prior-Guided Adaptive (PGA) variant achieved the highest performance
across all three reported metrics, with an AUPRC of 25.17, an AUROC of
69.49, and a PPV of 14.95. Compared with \textsc{OODQwen3-1.7B}, PGA
increased AUPRC by 0.70 percentage points, AUROC by 0.93 points, and PPV
by 2.61 points.
Diagnosis-specific continued pretraining improved performance for both
evaluated model architectures. Compared with ClinicalMamba-2.8B,
\textsc{OODMamba-2.8B} increased AUPRC from 10.35 to 11.15, AUROC from
65.22 to 66.11, and PPV from 10.92 to 11.74. Similarly,
\textsc{OODQwen3-1.7B} increased AUPRC from 20.79 to 24.47 and AUROC
from 67.59 to 68.56 relative to directly fine-tuned Qwen3-1.7B.

\subsection{Risk-Stratified Performance and Clinical Utility}

We evaluated risk-stratified performance among patients ranked in the
top 1\%, 2\%, 5\%, and 10\% of predicted 180-day opioid overdose risk
(Table~\ref{tab:risk_stratified}). The top-5\% threshold was prespecified
as the primary operating point. At this threshold,
\textsc{OverdoseMoE-PGA} achieved the highest PPV of 25.38 and recall of
21.14. By comparison, the strongest directly fine-tuned language-model
comparator, Qwen3-1.7B, achieved a PPV of 24.36 and recall of 20.29,
corresponding to absolute differences of 1.02 and 0.85 percentage points,
respectively. Other directly fine-tuned language models achieved lower
top-5\% PPV and recall. Qwen3-4B achieved a PPV of 22.58 and recall of
18.81, while Qwen3-8B achieved a PPV of 23.35 and recall of 19.45.
At the same top-5\% threshold, the PPV of 25.38 for
\textsc{OverdoseMoE-PGA} corresponded to a 4.22-fold risk enrichment
relative to the overdose prevalence of 6.01 in the held-out VHA test
cohort and a number needed to review of 3.9. In comparison, the
corresponding risk enrichment was 4.05-fold for Qwen3-1.7B, 3.76-fold
for Qwen3-4B, 3.89-fold for Qwen3-8B, and 3.67-fold for
\textsc{OODQwen3-1.7B}.
At other operating points, performance varied across models. At the
top-1\% threshold, \textsc{OODQwen3-1.7B} and
\textsc{OverdoseMoE-PGA} both achieved a PPV of 79.74 and recall of
13.31. At the top-2\% threshold, \textsc{OverdoseMoE-GLA} achieved the
highest PPV of 44.93 and recall of 15.01. Finally, at the top-10\%
threshold, \textsc{OverdoseMoE-EW} achieved the highest PPV of 17.78.

\begin{table*}[ht]
\centering
\scriptsize
\setlength{\tabcolsep}{4pt}
\renewcommand{\arraystretch}{1.12}
\resizebox{\textwidth}{!}{
\begin{tabular}{lcccccccccc}
\toprule
\textbf{Model}
& \multicolumn{4}{c}{\textbf{PPV @ Top-$k$\%}}
& \multicolumn{4}{c}{\textbf{Recall @ Top-$k$\%}}
& \textbf{Top-5\%} & \textbf{Top-5\%} \\
\cmidrule(lr){2-5}
\cmidrule(lr){6-9}
\cmidrule(lr){10-11}
& \textbf{1\%}
& \textbf{2\%}
& \underline{\textbf{5\%}}
& \textbf{10\%}
& \textbf{1\%}
& \textbf{2\%}
& \underline{\textbf{5\%}}
& \textbf{10\%}
& \textbf{Risk Enrichment}
& \textbf{NNR} \\
\midrule

BioMistral-7B
& 8.86 & 6.96 & 6.34 & 5.20
& 1.47 & 2.32 & 5.28 & 8.66
& 1.05$\times$ & 15.8 \\

Qwen2.5-3B
& 7.59 & 6.32 & 6.85 & 6.48
& 1.26 & 2.11 & 5.70 & 10.78
& 1.14$\times$ & 14.6 \\

ClinicalMamba-2.8B
& 10.12 & 8.22 & 7.61 & 6.48
& 1.69 & 2.74 & 6.34 & 10.78
& 1.27$\times$ & 13.1 \\

Qwen3-1.7B
& 58.22 & 40.50 & 24.36 & 18.55
& 9.72 & 13.53 & 20.29 & \textbf{30.86}
& 4.05$\times$ & 4.1 \\

Qwen3-4B
& 67.08 & 41.77 & 22.58 & 16.51
& 11.20 & 13.95 & 18.81 & 27.48
& 3.76$\times$ & 4.4 \\

Qwen3-8B
& 62.02 & 40.50 & 23.35 & 17.15
& 10.35 & 13.53 & 19.45 & 28.54
& 3.89$\times$ & 4.3 \\

\midrule

\textsc{OODMamba-2.8B}
& 17.72 & 17.08 & 15.22 & 14.36
& 2.95 & 5.70 & 12.68 & 23.89
& 2.53$\times$ & 6.6 \\

\textsc{OODQwen3-1.7B}
& \textbf{79.74} & 43.67 & 22.08 & 16.77
& \textbf{13.31} & 14.58 & 18.39 & 27.90
& 3.67$\times$ & 4.5 \\

\midrule

\textsc{OverdoseMoE-EW}
& 77.21 & 44.30 & 24.36 & \textbf{17.78}
& 12.89 & 14.79 & 20.29 & 29.59
& 4.05$\times$ & 4.1 \\

\textsc{OverdoseMoE-PGA}
& \textbf{79.74} & 42.40 & \textbf{25.38} & 17.66
& \textbf{13.31} & 14.16 & \textbf{21.14} & 29.38
& \textbf{4.22$\times$} & \textbf{3.9} \\

\textsc{OverdoseMoE-GLA}
& 78.48 & \textbf{44.93} & 24.87 & 17.40
& 13.10 & \textbf{15.01} & 20.72 & 28.96
& 4.14$\times$ & 4.0 \\

\bottomrule
\end{tabular}
}

\caption{
Risk-stratified PPV and recall for the evaluated models among patients
ranked in the top 1\%, 2\%, 5\%, and 10\% of predicted 180-day
opioid-overdose risk. \textsc{OverdoseMoE} includes three expert-fusion
variants: EW (Equal-Weight Fusion), PGA (Prior-Guided Adaptive Fusion),
and GLA (Global-Local Adaptive Fusion). The top-5\% threshold is
underlined because it was prespecified for targeted risk review.
Risk enrichment is calculated as PPV at the top-5\% threshold divided
by the overall opioid-overdose prevalence of 6.01\%. Number needed to
review (NNR) is calculated as 100 divided by PPV at the top-5\%
threshold. All PPV and recall values are reported as percentages.
}
\label{tab:risk_stratified}
\end{table*}

\subsection{Subgroup evaluation}

We evaluated \textsc{OODQwen3-1.7B} and the three
\textsc{OverdoseMoE} variants across age, sex, and race/ethnicity
subgroups in the held-out VHA test set
(Table~\ref{tab:moe_demographic_subgroups}). Excluding the
``Other or unknown'' race/ethnicity subgroup, PGA numerically improved
PPV and AUPRC relative to \textsc{OODQwen3-1.7B} in 9 of 11 subgroups
and AUROC in 10 of 11 subgroups.
Across age groups, the largest improvements were observed in several
middle-aged and older subgroups. Among patients aged 61--70 years, PGA
increased PPV from 14.25 to 15.96, AUPRC from 24.44 to 27.36,
and AUROC from 69.19 to 72.33 relative to
\textsc{OODQwen3-1.7B}. Among patients aged 71--80 years, GLA achieved
the highest AUROC of 62.79, compared with 59.19 for
\textsc{OODQwen3-1.7B}. Among patients aged 80 years or older, EW
achieved the highest AUROC of 66.93, compared with 53.63 for
\textsc{OODQwen3-1.7B}.
Among female patients, PGA achieved the highest PPV of 11.53, AUPRC of
28.71, and AUROC of 69.41. Among male patients, PGA achieved the highest
PPV of 13.45 and AUPRC of 23.75, whereas EW achieved the highest AUROC
of 68.22.
Among patients categorized as not Hispanic or Latino, PGA achieved the
highest PPV of 13.16 and AUPRC of 23.49, whereas EW achieved the highest
AUROC of 67.70. Among Hispanic or Latino patients, PGA achieved the
highest PPV of 16.98 and AUROC of 77.23, while
\textsc{OODQwen3-1.7B} achieved the highest AUPRC of 36.27.
Performance in the ``Other or unknown'' subgroup varied substantially
across models.

\begin{table*}[ht]
\centering
\scriptsize
\renewcommand{\arraystretch}{1.15}
\setlength{\tabcolsep}{3.5pt}

\resizebox{\textwidth}{!}{
\begin{tabular}{ll|ccc|ccc|ccc|ccc}
\toprule
& &
\multicolumn{3}{c|}{\textbf{\textsc{OODQwen3-1.7B}}} &
\multicolumn{3}{c|}{\textbf{\textsc{OverdoseMoE-EW}}} &
\multicolumn{3}{c|}{\textbf{\textsc{OverdoseMoE-PGA}}} &
\multicolumn{3}{c}{\textbf{\textsc{OverdoseMoE-GLA}}} \\
\cmidrule(lr){3-5}
\cmidrule(lr){6-8}
\cmidrule(lr){9-11}
\cmidrule(lr){12-14}

\textbf{Group} &
\textbf{Subgroup} &
\textbf{PPV} & \textbf{AUPRC} & \textbf{AUROC} &
\textbf{PPV} & \textbf{AUPRC} & \textbf{AUROC} &
\textbf{PPV} & \textbf{AUPRC} & \textbf{AUROC} &
\textbf{PPV} & \textbf{AUPRC} & \textbf{AUROC} \\
\midrule

\multirow{7}{*}{Age}
& 18--30
& 13.33 & 25.45 & 56.52
& 13.12 & 23.03 & 62.14
& 12.68 & 23.50 & 59.78
& 12.82 & 22.87 & 61.92 \\

& 31--40
& 11.89 & 21.73 & 64.32
& 11.42 & 20.59 & 64.64
& 12.53 & 22.07 & 65.65
& 11.16 & 20.43 & 64.57 \\

& 41--50
& 7.73 & 14.86 & 65.08
& 7.46 & 12.76 & 63.66
& 7.92 & 16.56 & 62.17
& 7.53 & 15.03 & 63.46 \\

& 51--60
& 12.53 & 22.42 & 67.70
& 13.11 & 23.84 & 68.27
& 13.23 & 23.44 & 68.44
& 12.81 & 23.54 & 68.09 \\

& 61--70
& 14.25 & 24.44 & 69.19
& 13.93 & 26.56 & 72.25
& 15.96 & 27.36 & 72.33
& 13.90 & 26.44 & 71.84 \\

& 71--80
& 13.18 & 26.04 & 59.19
& 12.50 & 26.72 & 62.54
& 14.86 & 26.18 & 62.31
& 13.15 & 26.92 & 62.79 \\

& 80+
& 16.66 & 27.78 & 53.63
& 14.28 & 29.57 & 66.93
& 16.66 & 28.38 & 58.26
& 12.50 &29.07 & 62.09 \\

\midrule

\multirow{2}{*}{Sex}
& Female
& 9.09 & 23.14 & 66.69
& 10.65 & 25.17 & 67.52
& 11.53 & 28.71 & 69.41
& 9.83 & 25.10 & 66.51 \\

& Male
& 12.73 & 22.61 & 66.10
& 12.44 & 23.52 & 68.22
& 13.45 & 23.75 & 68.01
& 12.36 & 23.33 & 68.01 \\

\midrule

\multirow{3}{*}{\makecell[l]{Race/\\ethnicity}}
& Not Hispanic or Latino
& 12.22 & 21.92 & 65.59
& 12.44 & 23.40 & 67.70
& 13.16 & 23.49 & 67.50
& 12.29 & 23.19 & 67.46 \\

& Hispanic or Latino
& 16.50 & 36.27 & 75.22
& 13.15 & 31.14 & 76.26
& 16.98 & 35.35 & 77.23
& 13.04 & 30.73 & 75.90 \\

& Other or unknown
& 5.00 & 20.00 & 96.39
& 0.00 & 3.33 & 73.87
& 4.76 & 8.33 & 90.09
& 0.00 & 3.22 & 72.97 \\
\bottomrule
\end{tabular}
}

\caption{
Subgroup performance of \textsc{OODQwen3-1.7B} and the three
\textsc{OverdoseMoE} variants for 180-day opioid overdose prediction
across age, sex, and race/ethnicity strata in the held-out VHA test set.
EW denotes Equal-Weight Fusion, PGA denotes Prior-Guided Adaptive Fusion,
and GLA denotes Global-Local Adaptive Fusion. Performance is reported as
PPV, AUPRC, and AUROC, with all values expressed as percentages.
Dashes indicate subgroup results that were not available or not reported.
Results for strata with very few positive events should be interpreted
descriptively because of the instability of subgroup-level estimates.
}
\label{tab:moe_demographic_subgroups}
\end{table*}

\subsection{External validation}

\begin{table*}[ht]
\centering
\scriptsize
\renewcommand{\arraystretch}{1.15}
\setlength{\tabcolsep}{5pt}

\begin{tabular}{llccc}
\toprule
\textbf{Family} & \textbf{Model} &
\textbf{AUPRC} & \textbf{AUROC} & \textbf{PPV} \\
\midrule

\multirow{7}{*}{Fine-tuned LM}
& Mamba-2.8B           & 10.96 & 55.08 & 7.50 \\
& BioMistral-7B        & 10.17 & 57.64 & 7.69 \\
& Qwen2.5-3B           & 7.27  & 50.12 & 7.24 \\
& ClinicalMamba-2.8B   & 8.50  & 48.78 & 7.00 \\
& Qwen3-1.7B           & 10.70 & 54.94 & 9.14 \\
& Qwen3-4B             & \textbf{14.11} & 58.94 & 10.01 \\
& Qwen3-8B             & 12.87 & 58.99 & 9.07 \\

\midrule

\multirow{2}{*}{Diagnosis-adapted}
& \textsc{OODMamba-2.8B} & 7.74  & 50.69 & 7.01 \\
& \textsc{OODQwen3-1.7B} & 12.43 & 58.94 & 9.29 \\

\midrule

\multirow{3}{*}{\textsc{OverdoseMoE}}
& \textbf{EW}  & 13.96 & 60.67 & 10.10 \\
& \textbf{PGA} & 13.07 & 60.68 & \textbf{10.24} \\
& \textbf{GLA} & 13.87 & \textbf{60.82} & 10.18 \\

\bottomrule
\end{tabular}

\caption{
External validation on an independent OUD cohort constructed from
MIMIC-IV for 180-day opioid overdose prediction.
\textsc{OverdoseMoE} includes three expert-fusion variants:
EW (Equal-Weight Fusion), PGA (Prior-Guided Adaptive Fusion), and
GLA (Global-Local Adaptive Fusion).
The best result for each metric is shown in bold.
}
\label{tab:model_transfer}

\end{table*}

To assess cross-cohort performance, we evaluated all models in an
independent OUD cohort constructed from MIMIC-IV
(Table~\ref{tab:model_transfer}). Among the individual models,
Qwen3-4B achieved the highest AUPRC of 14.11, whereas Qwen3-8B achieved
the highest AUROC of 58.99 among the directly fine-tuned language
models. The diagnosis-adapted \textsc{OODQwen3-1.7B} achieved an AUPRC
of 12.43, an AUROC of 58.94, and a PPV of 9.29.
Among the three \textsc{OverdoseMoE} variants,
\textsc{OverdoseMoE-GLA} achieved the highest AUROC of 60.82, while
\textsc{OverdoseMoE-PGA} achieved the highest PPV of 10.24.
\textsc{OverdoseMoE-EW} achieved the highest AUPRC of 13.96 among the
fusion variants, approaching the highest AUPRC of 14.11 observed across
all evaluated models with Qwen3-4B. All three \textsc{OverdoseMoE}
variants achieved higher AUPRC, AUROC, and PPV than
\textsc{OODQwen3-1.7B} in the external cohort.
Despite these improvements, absolute performance was generally lower
in MIMIC-IV than in the held-out VHA test cohort. For example,
\textsc{OverdoseMoE-PGA} decreased from an AUPRC of 25.17, an AUROC of
69.49, and a PPV of 14.95 in the VHA test cohort to an AUPRC of 13.07,
an AUROC of 60.68, and a PPV of 10.24 in MIMIC-IV. Similarly,
\textsc{OODQwen3-1.7B} decreased from an AUPRC of 24.47 and an AUROC of
68.56 in the VHA test cohort to an AUPRC of 12.43 and an AUROC of 58.94
in MIMIC-IV.
\subsection{Input Representation Ablation}
We further examined how different representations of longitudinal diagnosis
histories affected downstream opioid overdose prediction.
\textbf{Plain} represents the original chronological sequence of diagnosis
descriptions without additional structural or temporal information; for
example, ``Opioid dependence, uncomplicated; Essential hypertension; Major
depressive disorder.'' \textbf{Visit-tagged} groups diagnosis descriptions
recorded on the same date using explicit visit-boundary tokens; for example,
``<visit> Opioid dependence, uncomplicated; Essential hypertension
</visit> <visit> Major depressive disorder </visit>.''
\textbf{Time-augmented} appends the corresponding date to each diagnosis
description; for example, ``Opioid dependence, uncomplicated [2017-03-02];
Essential hypertension [2017-03-02]; Major depressive disorder
[2017-03-03].'' \textbf{ICD-only} removes diagnosis descriptions and retains
only the ICD codes in chronological order; for example,
``F11.20, I10, F32.9.''

\begin{figure*}[htp]
\centering
\includegraphics[width=\textwidth]{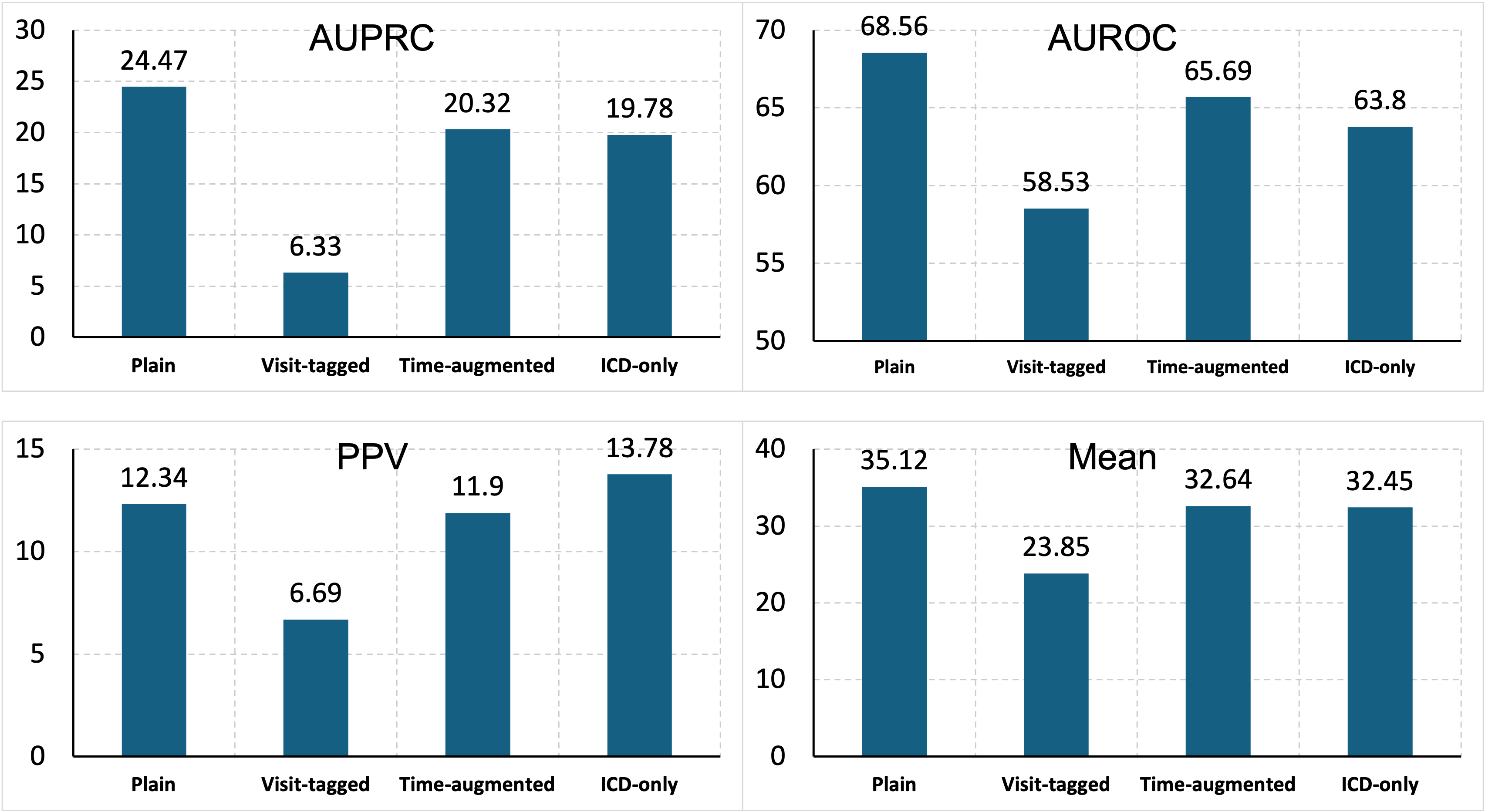}
\caption{\textbf{Ablation of Input Representations for OODQwen3-1.7B.}}
\label{fig:input_1_7b}
\end{figure*}

\begin{figure*}[htp]
\centering
\includegraphics[width=\textwidth]{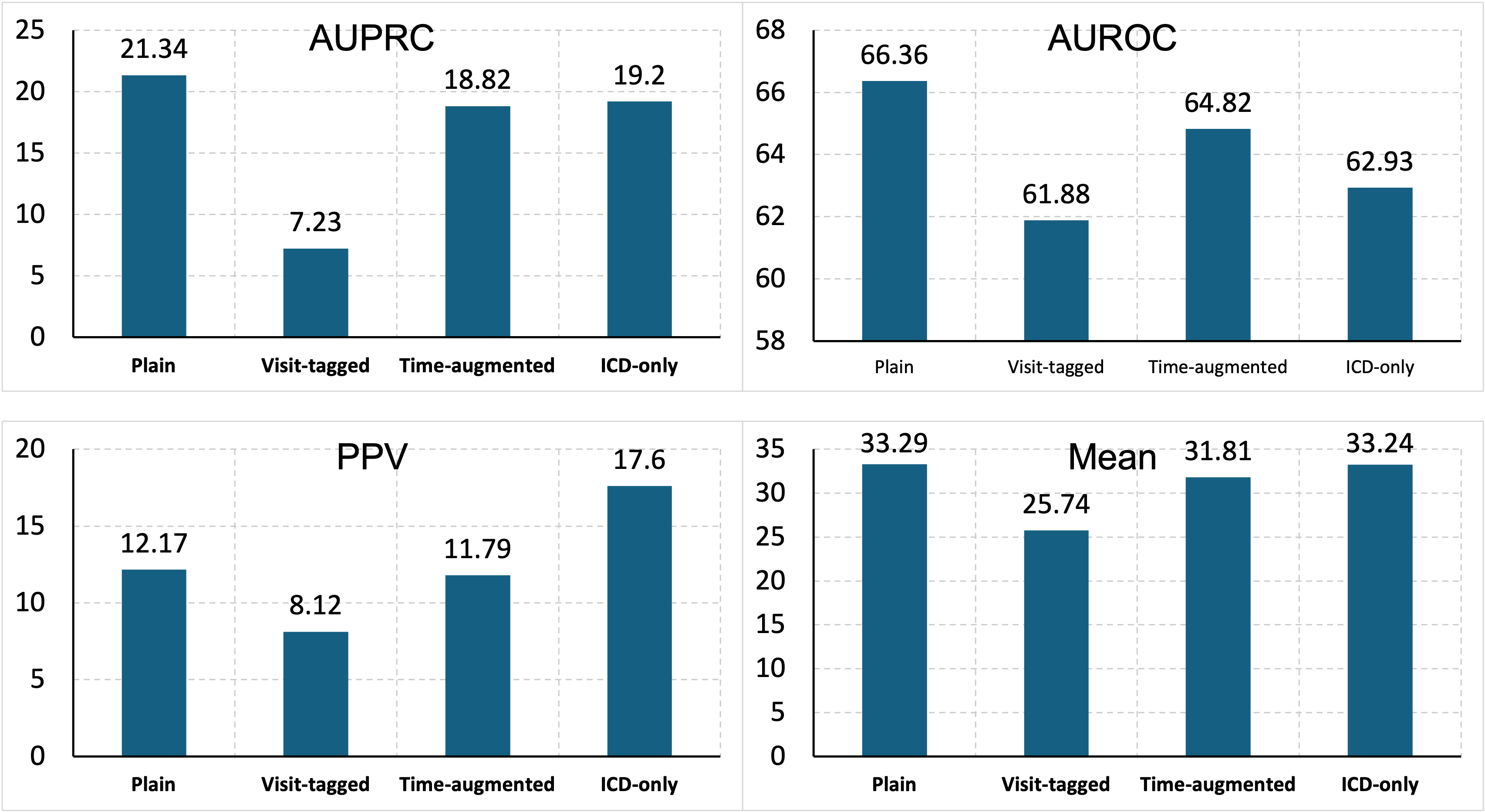}
\caption{\textbf{Ablation of Input Representations for Qwen3-4B.}}
\label{fig:input_4b}
\end{figure*}

\begin{figure*}[htp]
\centering
\includegraphics[width=\textwidth]{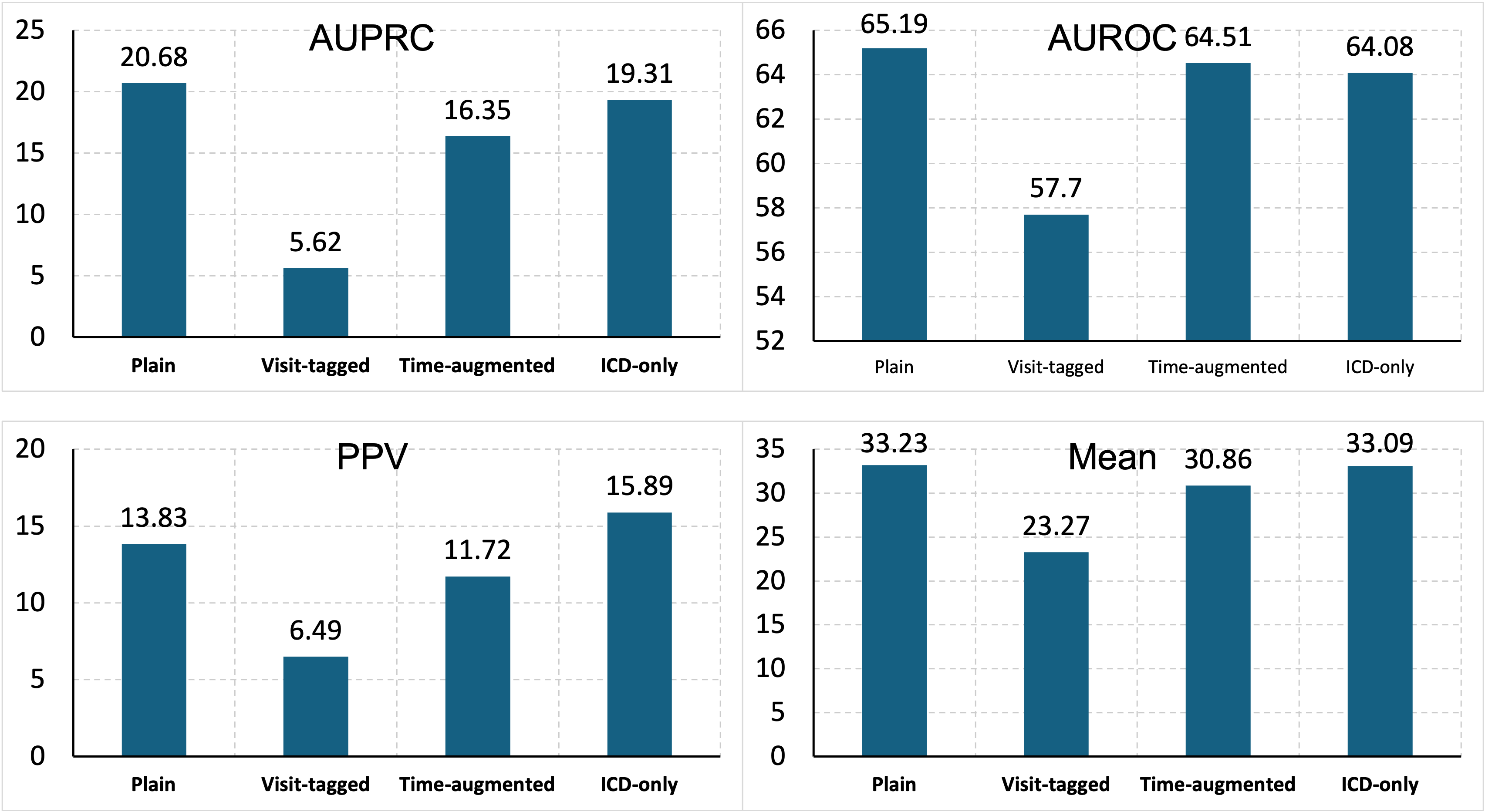}
\caption{\textbf{Ablation of Input Representations for OODQwen3-8B.}}
\label{fig:input_8b}
\end{figure*}

Across OODQwen3-1.7B, Qwen3-4B, and Qwen3-8B, the Plain representation
achieved the highest mean performance, whereas Visit-tagged input
consistently produced the lowest performance
(Figs.~\ref{fig:input_1_7b}--\ref{fig:input_8b}). For OODQwen3-1.7B,
the mean score decreased from 35.12 with Plain input to 23.85 with
Visit-tagged input. Corresponding differences were observed for Qwen3-4B
(33.29 vs. 25.74) and Qwen3-8B (33.23 vs. 23.27).
Time-augmented input achieved mean scores of 32.64, 31.81, and 30.86 for
OODQwen3-1.7B, Qwen3-4B, and Qwen3-8B, respectively, whereas ICD-only
input achieved corresponding mean scores of 32.45, 33.24, and 33.09.
Although Plain input achieved the highest mean performance across all
three model scales, ICD-only input achieved the highest PPV for Qwen3-4B
(17.60) and Qwen3-8B (15.89).

%% file: 4_Discussion.tex
\section{Discussion}

In this study, we evaluated diagnosis-specific continued pretraining and
multi-expert integration for predicting 180-day opioid overdose risk from
longitudinal diagnosis histories. Three main findings emerged. First,
diagnosis-specific continued pretraining improved predictive performance
relative to direct fine-tuning for both the Qwen- and Mamba-based
architectures, with the largest gain observed for \textsc{OODQwen3-1.7B}.
Second, integrating Qwen-based experts of different model scales through
\textsc{OverdoseMoE} further improved overall discrimination and positive
predictive value, with the PGA variant achieving the strongest overall
performance in the held-out VHA cohort. Third, these improvements extended
to clinically relevant high-risk operating points and were generally
maintained across demographic subgroups and in an independent MIMIC-IV
cohort, although the magnitude of benefit varied across populations and
evaluation settings.

The improvement associated with diagnosis-specific continued pretraining
suggests that adaptation to longitudinal clinical diagnosis sequences can
provide information beyond conventional task-specific fine-tuning. Unlike
general-purpose pretraining, continued exposure to longitudinal diagnosis
histories may allow the model to better represent recurring comorbidities,
disease progression, and combinations of diagnoses that precede subsequent
overdose. The improvement was observed for both
\textsc{OODQwen3-1.7B} and \textsc{OODMamba-2.8B}, indicating that the
benefit was not restricted to a single neural architecture. However, the
substantially larger gain for \textsc{OODQwen3-1.7B} also indicates that
the magnitude of benefit depends on the underlying model architecture and
its capacity to use the adapted longitudinal representation.

Beyond diagnosis-specific adaptation, the additional improvement obtained
with \textsc{OverdoseMoE} suggests that experts of different model scales
capture partially complementary predictive information. Models with
different capacities may differ in the longitudinal patterns to which they
are most sensitive and in the errors they make for individual patients.
Combining their predictions therefore provides an opportunity to reduce
dependence on any single representation. Among the evaluated fusion
strategies, PGA provided the strongest overall performance in the primary
VHA analysis and showed relatively consistent gains across many demographic
subgroups. This pattern suggests that combining a prior preference over
experts with patient-specific adaptive weighting may provide a useful
balance between stable global performance and sample-level flexibility.
At the same time, no fusion strategy was uniformly superior at every risk
threshold or in every subgroup, indicating that the relative contribution
of individual experts can vary across operating points and patient
populations.

The risk-stratified analysis further illustrates the potential clinical
relevance of these improvements. At the prespecified top-5\% operating
point, \textsc{OverdoseMoE-PGA} achieved a PPV of 25.38 and a recall of
21.14, corresponding to a 4.22-fold enrichment over the 6.01 overdose
prevalence in the held-out VHA cohort and a number needed to review of
3.9. In a setting where clinical resources for overdose prevention are
limited, concentrating future overdose events within a relatively small
high-risk group could facilitate prioritization for additional assessment
or preventive services. Nevertheless, these findings reflect retrospective
risk stratification rather than the effect of a deployed clinical
intervention. Prospective evaluation would therefore be required to
determine whether model-guided prioritization improves clinical outcomes or
resource allocation in practice.

The subgroup analyses showed that the benefit of multi-expert fusion was
not confined to a single demographic stratum, but the magnitude and type of
improvement varied across age, sex, and race/ethnicity groups. PGA showed
the most frequent numerical improvements relative to
\textsc{OODQwen3-1.7B}, whereas EW and GLA achieved the highest
discrimination in several individual strata. Such variation is consistent
with heterogeneity in the predictive patterns represented across experts
and provides further motivation for adaptive rather than fixed expert
weighting. Importantly, subgroup performance should not be interpreted as
evidence of demographic fairness. Estimates in smaller strata, particularly
the ``Other or unknown'' race/ethnicity group, were unstable, and larger
samples with sufficient outcome events will be needed to characterize
performance differences more reliably across patient populations.

External validation in MIMIC-IV provided a more challenging evaluation
setting. Absolute performance was lower for most models than in the
held-out VHA cohort. This reduction may reflect differences between the two
healthcare systems in demographic composition, clinical setting,
documentation and coding practices, and outcome prevalence. In particular,
the prevalence of 180-day opioid overdose was 1.38\% in MIMIC-IV compared
with approximately 6\% across the VHA cohorts. Such differences can alter
both the distribution of longitudinal diagnosis histories and their
association with subsequent overdose, thereby limiting the transportability
of models developed within a single healthcare system. Despite this shift,
the three \textsc{OverdoseMoE} variants remained among the strongest
approaches in the external cohort, with GLA achieving the highest AUROC and
overall average score and PGA achieving the highest PPV. One potential
explanation is that integrating experts of different scales reduces
dependence on errors or representations specific to any single model,
providing greater stability under population shift. However, validation in
a single external cohort does not establish broad transportability, and
evaluation across additional healthcare systems and patient populations is
needed.

The input-representation ablation provides additional insight into how
longitudinal diagnosis information is used by the models. Plain input
achieved the highest mean performance across the three evaluated model
scales, whereas explicit Visit-tagged representations consistently produced
the lowest performance. The reduction with Visit-tagged input may reflect
the limited fidelity of grouping diagnoses by calendar date as a proxy for
a true clinical encounter. Diagnoses recorded on the same date can arise
from different clinical activities or documentation processes, and explicit
visit-boundary tokens may therefore impose structure that does not
correspond closely to clinically meaningful encounters. Repeated boundary
tokens also increase sequence length and may introduce additional
representational complexity without providing information directly relevant
to subsequent overdose risk.

In contrast, ICD-only representations remained competitive, particularly
for the larger Qwen models, and produced the highest PPV for Qwen3-4B and
Qwen3-8B. This finding indicates that diagnosis identity and chronological
ordering themselves contain a substantial proportion of the predictive
signal. Explicitly adding calendar dates produced only modest changes in
performance relative to the Plain representation, suggesting that detailed
calendar-time information did not provide additional benefit beyond the
original chronological ordering in the evaluated setting. Importantly,
the poorer performance of Visit-tagged input pertains to explicit
visit-boundary encoding during downstream prediction and should not be
interpreted as evidence that encounter structure is uninformative during
diagnosis-specific continued pretraining.

Taken together, these findings indicate that the combination of
diagnosis-specific adaptation and multi-expert integration can improve
opioid overdose risk prediction from longitudinal diagnosis histories.
The results also highlight that model performance depends not only on model
scale but also on how longitudinal clinical information is represented and
combined across predictors. Further validation across healthcare systems,
prospective evaluation of risk-guided interventions, and assessment of
performance across adequately powered demographic subgroups will be
important before clinical deployment.

%% file: 5_method.tex
\section{Methods}

\subsection{Data sources and ethical approval}

This study used longitudinal electronic health record (EHR) data from the Veterans Health Administration (VHA) Corporate Data Warehouse (CDW), the clinical data system of the largest integrated health care network in the United States, comprising more than 1{,}200 medical centers and clinics. For cross-cohort transfer evaluation, we additionally constructed an independent opioid use disorder (OUD) cohort from MIMIC-IV, a publicly available database of de-identified patient records from a single academic tertiary-care center~\cite{johnson2023mimic}. This study was approved by the Institutional Review Board of the US Veterans Affairs (VA) Bedford Health Care and conducted in accordance with the principles of the Declaration of Helsinki. A waiver of informed consent was obtained due to minimal risk to participants (IRB Net ID: 1603552-28).

\subsection{Study design, prediction task, and outcome definition}

The study was a retrospective longitudinal prognostic analysis of 180-day opioid overdose risk among VHA patients with a first recorded diagnosis of opioid use disorder. For each patient, the index date was defined as the first recorded OUD diagnosis. The model input was the 12 months of diagnostic history preceding the index date, and the prediction target was an opioid overdose event within the subsequent 180 days. The 180-day prediction horizon was chosen to align with VHA-level surveillance and care-coordination cadences used after an OUD diagnosis~\cite{oliva2017development}.
Opioid overdose was ascertained using the ICD-based mapping of Glanz et al.~\cite{glanz2019association}. Patients with at least one qualifying opioid-overdose diagnosis code during the 180-day prediction window were labeled positive. Patients without a qualifying opioid-overdose diagnosis code during the prediction window were labeled negative.

\subsection{Model framework}
\begin{figure*}[htp]
\centering
\includegraphics[width=\textwidth]{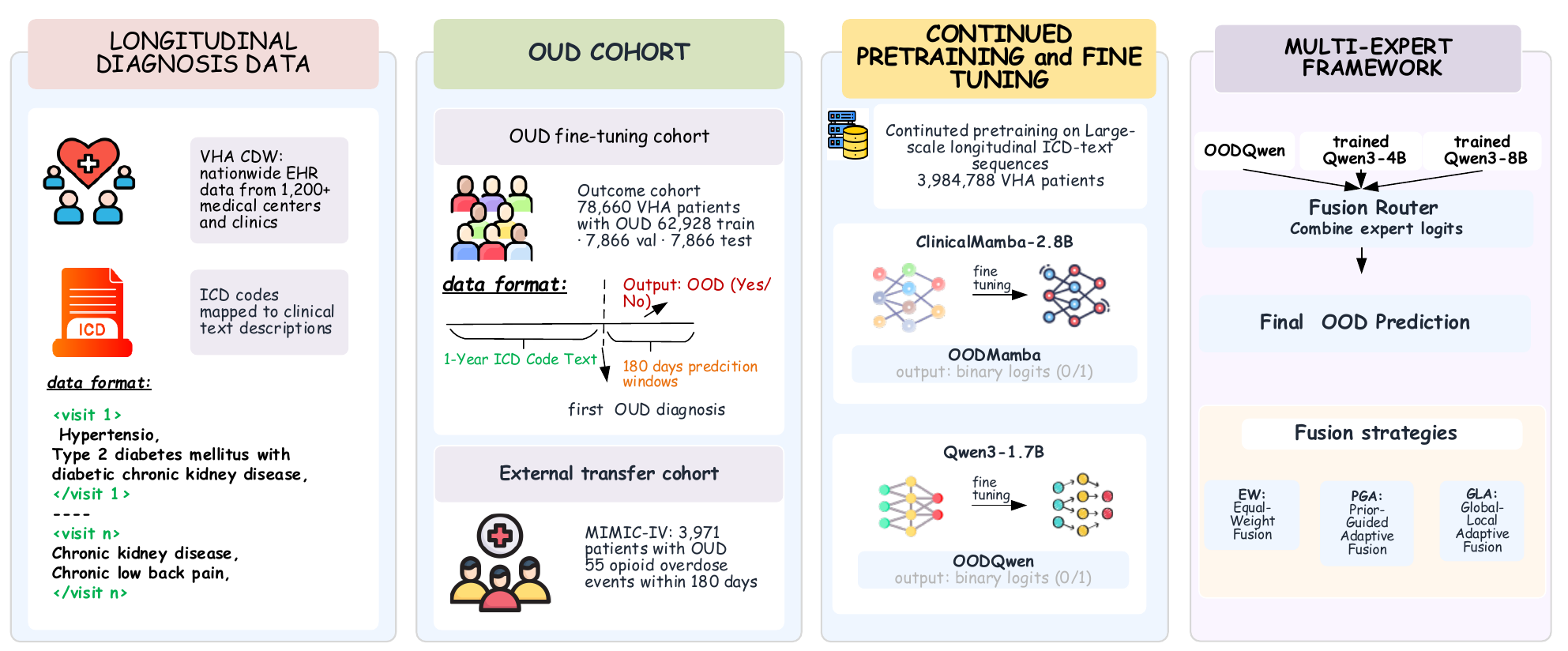}
\caption{
\textbf{Overview of the OverdoseMoE framework for 180-day opioid overdose risk prediction.}
The framework includes four components.
(1) \textit{Longitudinal diagnosis data}: nationwide VHA electronic health record data are converted into longitudinal sequences of clinical diagnosis descriptions for continued pretraining.
(2) \textit{Cohort construction}: an OUD cohort is used for model fine-tuning and evaluation, with one-year longitudinal diagnosis histories as input and opioid overdose within 180 days as the prediction outcome; an independent MIMIC-IV OUD cohort is used for external evaluation.
(3) \textit{Continued pretraining and task-specific adaptation}: ClinicalMamba-2.8B and Qwen3-1.7B are continued pretrained on large-scale VHA diagnosis sequences to obtain OODMamba and OODQwen, followed by fine-tuning for binary opioid overdose prediction.
(4) \textit{Multi-expert fusion}: OODQwen and independently fine-tuned Qwen3-4B and Qwen3-8B models provide complementary prediction logits, which are combined using equal-weight (EW), prior-guided adaptive (PGA), or global--local adaptive (GLA) fusion to produce the final overdose risk prediction.
}
\label{fig:wholeflow}
\end{figure*}

We developed OverdoseMoE as a unified framework for 180-day opioid
overdose risk prediction from longitudinal diagnosis histories. The
framework integrates longitudinal clinical representation, domain
adaptation, task-specific prediction, and multi-expert fusion. First,
diagnosis records are converted into chronologically ordered clinical
text sequences and used to construct the pretraining, OUD, and external
evaluation cohorts. ClinicalMamba-2.8B and Qwen3-1.7B are then adapted
to the VHA domain through continued pretraining and further optimized
for binary overdose prediction. Finally, predictions from complementary
Qwen-based experts are combined using three fusion strategies to produce
the final patient-level risk estimate. Figure~\ref{fig:wholeflow}
summarizes the overall workflow.
% \subsection{VHA cohorts}

\subsubsection{Data  and Cohort Construction}
\paragraph{Longitudinal diagnosis data (pretraining cohort).}
The pretraining cohort included 3{,}984{,}788 unique patients who
received care at more than 1{,}200 VHA facilities between January 1,
2016 and December 31, 2019. Of these patients, 3{,}944{,}418 were used
for continued pretraining and 40{,}370 were reserved for pretraining
validation. For each patient, diagnosis codes recorded across
longitudinal visits were mapped to their corresponding clinical text
descriptions. The mapped diagnoses were then organized by visit and
concatenated in chronological order to form a longitudinal diagnosis
sequence. Examples of mapped diagnoses include \textit{essential
hypertension}, \textit{type 2 diabetes mellitus}, and \textit{low back
pain}. An example of the resulting input representation is provided in
Supplementary Note~\ref{inputwxample}.

\paragraph{OUD fine-tuning cohort.}
We constructed a downstream OUD cohort consisting of 78{,}660 VHA
patients with an initial OUD diagnosis recorded between January 1, 2017
and June 1, 2019. Patients were partitioned at the patient level into
training, validation, and test sets containing 62{,}928, 7{,}866, and
7{,}866 patients, respectively. For each patient, the diagnostic history
from the 12 months preceding the index date was used as model input.
Patients with fewer than 12 months of diagnostic history before the
index date were excluded. No imputation was applied within the retained
observation window. Outcome counts across splits are summarized in
Table~\ref{tab:dataset_stats}, and demographic characteristics are
summarized in Table~\ref{tab:demographics}.

\paragraph{External evaluation cohort.}
We constructed an independent external evaluation cohort from MIMIC-IV
to assess cross-cohort generalizability. The cohort included 3{,}971
patients with OUD, of whom 55 experienced an opioid overdose within
180 days. The same prediction task and longitudinal diagnosis
representation used in the VHA cohort were applied to the MIMIC-IV
cohort. No MIMIC-IV data were used for model training or model
selection. Cohort size and 180-day overdose prevalence are summarized
in Table~\ref{tab:dataset_stats}, and demographic characteristics are
reported in Table~\ref{tab:demographics}.

% Figure~\ref{fig:wholeflow} summarizes the overall OverdoseMoE framework, which integrates domain adaptation, task-specific prediction, and multi-expert fusion for 180-day opioid overdose risk prediction.

\subsubsection{Continued pretraining and task-specific fine-tuning}

We developed two domain-adapted models, \textsc{OODMamba} and
\textsc{OODQwen}, based on ClinicalMamba-2.8B and Qwen3-1.7B,
respectively. ClinicalMamba-2.8B uses the Mamba selective state-space
architecture~\cite{gu2024mamba}, whereas Qwen3-1.7B uses a
transformer-based architecture. Both models followed the same two-stage
adaptation pipeline. First, the backbone models underwent continued
pretraining on large-scale longitudinal VHA diagnosis sequences.
Second, the adapted models were fine-tuned for 180-day opioid overdose
prediction using the VHA OUD cohort.

\paragraph{Continued pretraining.}
Chronologically ordered diagnosis-text sequences from the VHA
pretraining cohort were used to adapt both backbone models to the VHA
clinical domain. Continued pretraining used an autoregressive next-token
prediction objective,

\begin{equation}
\mathcal{L}_{\mathrm{CPT}}
=
-\sum_{t=1}^{T}
\log
P(x_t \mid x_{<t}),
\label{eq:cpt_objective}
\end{equation}

where \(x_t\) denotes the token at position \(t\) in a longitudinal
diagnosis sequence. This stage enabled the models to learn the clinical
terminology, temporal ordering, and longitudinal patterns present in VHA
diagnosis histories. Continued pretraining of ClinicalMamba-2.8B
produced \textsc{OODMamba}, whereas continued pretraining of
Qwen3-1.7B produced \textsc{OODQwen}.

\paragraph{Task-specific fine-tuning.}
Following continued pretraining, both domain-adapted models were
fine-tuned for 180-day opioid overdose prediction. For each patient, the
one-year longitudinal diagnosis history preceding the index date was
used as input. The prediction task was formulated as binary
classification,

\begin{equation}
y(x)\in\{0,1\},
\end{equation}

where \(y=1\) denotes an opioid overdose during the 180-day prediction
window and \(y=0\) denotes no overdose.
For \textsc{OODMamba}, a binary classification head was added to the
domain-adapted backbone and optimized using binary cross-entropy loss.
Fine-tuning used AdamW with a learning rate of \(2\times10^{-4}\), a
batch size of 16, a maximum sequence length of 2,048 tokens, weight
decay of 0.01, and dropout of 0.1 in the classification head. The model
was trained for up to 20 epochs, and the checkpoint with the lowest
validation loss was retained for evaluation. Class-weighted sampling
was used to address outcome imbalance.
For \textsc{OODQwen}, parameter-efficient fine-tuning was performed
using low-rank adaptation (LoRA). LoRA adapters were applied to the
query and value projection matrices (\texttt{q\_proj} and
\texttt{v\_proj}) with rank \(r=16\), scaling parameter
\(\alpha_{\mathrm{LoRA}}=16\), and dropout of 0.1. The maximum input
length was 2,000 tokens, and optimization used
\textit{BCEWithLogitsLoss}. The resulting model produced logits for the
non-overdose and overdose classes. These logits were used for
individual-model evaluation and as inputs to the subsequent
multi-expert fusion framework.
All training was performed on NVIDIA A100 80GB GPUs.

\subsubsection{Multi-expert fusion}

We developed OverdoseMoE to integrate complementary predictions from
three Qwen-based experts: \textsc{OODQwen}, Qwen3-4B, and Qwen3-8B.
All experts received the same longitudinal diagnosis representation and
were optimized for the same 180-day opioid overdose prediction task,
but differed in model capacity and learned representations.
For patient \(x\), expert \(i\in\{1,2,3\}\) produces a two-class logit
vector \(\mathbf{z}_i(x)\) and the corresponding class probabilities,

\begin{equation}
\mathbf{z}_i(x)=[z_{i0}(x),z_{i1}(x)],
\qquad
\mathbf{p}_i(x)=\operatorname{softmax}(\mathbf{z}_i(x)).
\label{eq:expert_outputs}
\end{equation}
The three experts are combined at the logit level using

\begin{equation}
\mathbf{z}_{\mathrm{fuse}}(x)
=
\sum_{i=1}^{3} w_i(x)\mathbf{z}_i(x),
\qquad
\sum_{i=1}^{3}w_i(x)=1,
\label{eq:general_fusion}
\end{equation}
where \(w_i(x)\) denotes the contribution of expert \(i\) for patient
\(x\). We evaluated three strategies for determining these weights.

\paragraph{Equal-weight fusion.}
Equal-weight fusion (EW) provides a parameter-free ensemble in which
all experts contribute equally,

\begin{equation}
w_i^{\mathrm{EW}}=\frac{1}{3},
\qquad
\mathbf{z}_{\mathrm{EW}}(x)
=
\frac{1}{3}\sum_{i=1}^{3}\mathbf{z}_i(x).
\label{eq:ew_fusion}
\end{equation}

\paragraph{Prior-guided adaptive fusion.}
Prior-guided adaptive fusion (PGA) combines validation-derived expert
priors with patient-specific predictive confidence. The global prior weights were predefined as
\(\boldsymbol{\pi}=(0.30,\,0.50,\,0.20)\), with greater weight assigned
to experts showing stronger individual performance on the validation
cohort. Accordingly, \textsc{OODQwen} received the largest prior weight.
For each expert, predictive entropy was used to quantify uncertainty,
and lower entropy was assigned greater patient-specific weight,
\begin{equation}
H_i(x)
=
-\sum_{c\in\{0,1\}}p_{ic}(x)\log p_{ic}(x),
\qquad
a_i(x)
=
\frac{\exp[-\beta H_i(x)]}
{\sum_{j=1}^{3}\exp[-\beta H_j(x)]}.
\label{eq:pga_entropy}
\end{equation}
The final PGA weight interpolates between the global prior and the
patient-specific confidence weight,
\begin{equation}
w_i^{\mathrm{PGA}}(x)
=
(1-\lambda)\pi_i+\lambda a_i(x),
\qquad
\mathbf{z}_{\mathrm{PGA}}(x)
=
\sum_{i=1}^{3}w_i^{\mathrm{PGA}}(x)\mathbf{z}_i(x),
\label{eq:pga_fusion}
\end{equation}
with \(\beta=1\). The mixing coefficient \(\lambda\) was selected on
the validation cohort by maximizing AUPRC over
\(\{0,\,0.01,\,0.02,\,0.05,\,0.10,\,0.20,\,0.30\}\).
When multiple values achieved the same validation performance, the
smaller value was retained. Thus, \(\lambda=0\) corresponds to fixed
prior-weighted fusion, whereas larger values allow progressively greater
patient-specific adaptation.

\paragraph{Global--local adaptive fusion.}
Global--local adaptive fusion (GLA) directly combines population-level
expert performance with patient-specific predictive uncertainty. Global
expert quality was defined by validation-set AUPRC.
% \begin{equation}
% Q_i=\operatorname{AUPRC}^{\mathrm{val}}_i,
% \qquad
% (Q_1,Q_2,Q_3)=(0.2185,\,0.1792,\,0.2068).
% \label{eq:gla_quality}
% \end{equation}
For each patient, the global quality score and predictive entropy were
combined into a routing score and normalized across experts,
\begin{equation}
s_i(x)=\alpha Q_i-\beta H_i(x),
\qquad
w_i^{\mathrm{GLA}}(x)
=
\frac{\exp[s_i(x)]}
{\sum_{j=1}^{3}\exp[s_j(x)]},
\label{eq:gla_weight}
\end{equation}
where \(\alpha=\beta=1\). The final prediction logits are
\begin{equation}
\mathbf{z}_{\mathrm{GLA}}(x)
=
\sum_{i=1}^{3}
w_i^{\mathrm{GLA}}(x)\mathbf{z}_i(x).
\label{eq:gla_fusion}
\end{equation}

GLA therefore assigns greater weight to experts with stronger
validation-set discrimination and lower uncertainty for the current
patient, whereas PGA adaptively adjusts predefined global prior weights
using patient-specific confidence.

\subsection{Baseline models}
To assess the contribution of sequence architecture, task-specific fine-tuning, diagnosis-domain adaptation, and multi-expert fusion, we compared the proposed models with baselines spanning four model families, consistent with the organization of the primary results table: deep-sequence models, fine-tuned language models, diagnosis-adapted models, and multi-expert fusion models.

The \textit{deep-sequence} baselines included a gated recurrent unit (GRU) network~\cite{cho2014gru} and Mamba-2.8B~\cite{gu2023mamba}, representing recurrent and selective state-space approaches for modeling chronologically ordered longitudinal diagnosis sequences.
The \textit{fine-tuned language-model} baselines included BioMistral-7B~\cite{labrak2024biomistral}, Qwen2.5-3B~\cite{yang2024qwen25}, ClinicalMamba-2.8B~\cite{yang2024clinicalmamba}, and three Qwen3 variants with 1.7B, 4B, and 8B parameters~\cite{yang2025qwen3}. All models were evaluated using the same patient-level data splits, one-year observation window, 180-day prediction horizon, and binary opioid-overdose outcome definition. Model-specific fine-tuning procedures followed the settings described above.
The \textit{diagnosis-adapted} models consisted of \textsc{OODMamba}-2.8B and \textsc{OODQwen}-1.7B. These models were obtained by continued pretraining of ClinicalMamba-2.8B and Qwen3-1.7B, respectively, on large-scale longitudinal VHA diagnosis histories before downstream fine-tuning for opioid-overdose prediction. ClinicalMamba-2.8B therefore served as the direct pretraining control for \textsc{OODMamba}, while Qwen3-1.7B served as the corresponding backbone control for \textsc{OODQwen}; these paired comparisons isolate the contribution of VHA diagnosis-specific continued pretraining.
Finally, the \textit{OverdoseMoE} family included the three proposed multi-expert fusion strategies: equal-weight fusion (EW), prior-guided adaptive fusion (PGA), and global--local adaptive fusion (GLA). All three approaches combined predictions from the same set of task-specific experts and differed only in how expert contributions were weighted, enabling direct evaluation of the benefit of adaptive expert routing.

% \textsc{OODMamba} was benchmarked against reference models spanning four families aligned with the primary results table: classical machine-learning models, deep-sequence models, zero-shot language-model baselines, and fine-tuned language-model baselines. Classical models included random forest, logistic regression, and support vector machine models operating on bag-of-ICD representations derived from the 12-month observation window. Deep-sequence baselines included GRU~\cite{chung2014empirical} and TransformerEHR~\cite{yang2023transformehr}, trained on chronologically ordered diagnosis-code sequences.

% Three publicly released language models were evaluated: Clinical Mamba-2.8B~\cite{yang2024clinicalmamba}, Qwen2.5-3B~\cite{qwen2.5}, BioMistral-7B~\cite{labrak2024biomistral}. These models were first evaluated in zero-shot form using task-specific prompts (Supplementary Note~2). Where applicable, they were then fine-tuned using the same binary classification head and optimization setup as \textsc{OODMamba}. Clinical Mamba-2.8B served as the primary pretraining control because it shares architecture and task setup with \textsc{OODMamba} but differs in pretraining corpus.

% , and BioLLaMA3-8B~\cite{ContactDoctor_MEDLLM}

\subsection{Evaluation and statistical analysis}

Predictive performance was evaluated on the held-out 7{,}866-patient VHA test set using positive predictive value (PPV), area under the precision-recall curve (AUPRC), and area under the receiver operating characteristic curve (AUROC). For the primary model-comparison table, binary predictions were obtained by taking the argmax over the binary classification logits, and PPV was computed from these binary predictions.

Because the intended use case was prioritization of a limited high-risk subgroup for prevention review, we also evaluated risk concentration at prespecified top-$k$ operating points. Patients were ranked by predicted opioid-overdose risk, and PPV and recall were computed among patients in the top 1\%, 2\%, 5\%, and 10\% of predicted risk. The top-5\% threshold was prespecified as the primary operating point because it represents a pragmatic registry-based review threshold: selective enough to define a manageable high-risk subgroup, while broad enough to capture more events than highly restrictive alert thresholds. This framing is consistent with prior work on EHR-integrated risk stratification and clinical implementation, in which high-risk strata are used to prioritize patients for review, outreach, or clinician notification under finite clinical resources~\cite{oliva2017development,lee2020clinical}. The 1\%, 2\%, and 10\% thresholds were included to examine more selective and broader review scenarios. For the top-5\% threshold, the observed-to-expected (O/E) ratio was calculated as PPV divided by the opioid-overdose prevalence in the held-out VHA test set, and number needed to review was calculated as the inverse of PPV.

Cross-cohort transfer was evaluated by applying the VHA-trained models to the MIMIC-IV OUD cohort using the same diagnosis-history representation and evaluation metrics. This analysis was interpreted as a cross-cohort transportability assessment rather than as evidence of broad generalizability. Exploratory descriptive subgroup analyses for the primary VHA test set were conducted across prespecified strata defined by age, sex, and race/ethnicity. Strata with fewer than 10 positive events were reported descriptively and were not used for subgroup comparisons.

\textsc{OODMamba} was evaluated as a risk-ranking tool at prespecified high-risk operating points. Use of model scores for absolute-risk estimation or threshold-based deployment would require dedicated calibration analysis on the target population and, if indicated, explicit recalibration.

%% file: 6_Other.tex
\section*{Data and Code Availability}

The data used in this study are derived from the U.S. Department of
Veterans Affairs and are not publicly available due to data-use and
privacy restrictions; approval by the Department of Veterans Affairs is
required for data access. MIMIC-IV~\cite{johnson2023mimic} is publicly
available under a PhysioNet credentialed data-use agreement. Cohort
construction scripts and prediction outputs for the MIMIC-IV transfer
analysis can be shared on request, subject to applicable data-use
agreements. Code associated with this study is available at
\url{https://github.com/ToneLi/OverdoseMoE/}.

% \section*{Code availability}
% Code for model pretraining, fine-tuning, and evaluation is available
% from the corresponding author on reasonable request. Key packages used
% include \texttt{mamba-ssm} for the Mamba backbone,
% \texttt{transformers} (Hugging Face) for language-model comparators,
% and \texttt{scikit-learn} for classical machine-learning baselines.
% Prompt templates used for zero-shot language-model comparators are
% provided in Supplementary Note~2.   %The code is available at \url{https://github.com/ToneLi/OOD_LLM_diagnosis}

\section*{Acknowledgements}

Research reported in this study was supported in part by research grants R01DA056470, R01AG080670, and 1I01HX003711. The content is solely the responsibility of the authors and does not necessarily represent the official views of the National Institutes of Health or the U.S. Department of Veterans Affairs, Veterans Health Administration.
% Recommended structure:
% This study was funded by [FUNDING AGENCY] under award number [GRANT
% NUMBER]. The funder had no role in study design, data collection,
% analysis, interpretation, or manuscript preparation. [Optional:
% acknowledge collaborators, data stewards, or computing-resource
% support.]

\section*{Author contributions}
% \textbf{Mingchen Li}, \textbf{Feiyun Ouyang}, and \textbf{Hong Yu} conceived and designed the study. 
Mingchen Li conducted the experiments, performed the analyses, and drafted the manuscript. Rohan Pandey conducted experiments and analyses, contributed to
  the study structure, and drafted the manuscript. Junhui Qian contributed to cohort construction. Feiyun Ouyang contributed to study design and provided
  medical guidance. Sunjae Kwon contributed to construction of the transfer evaluation cohort. Hong Yu supervised the study. All authors revised the manuscript, approved the final version, and agreed with the conclusions.

% Recommended structure using author initials:
% [INITIAL] and [INITIAL] conceived and designed the study. [INITIAL]
% conducted the experiments, performed the analyses, and drafted the
% manuscript. [INITIAL] contributed to [SPECIFIC CONTRIBUTION]. [INITIALS
% OF REMAINING AUTHORS] provided critical feedback and contributed to
% manuscript editing. [INITIAL] supervised the study. All authors revised
% the manuscript, approved the final version, and agreed with the
% conclusions.

\section*{Competing interests}

The authors declare no competing interests.

%% file: 8_appendix.tex
% Supplementary numbering
\section*{Supplementary Information}

\setcounter{figure}{0}
\setcounter{table}{0}
\renewcommand{\thefigure}{S\arabic{figure}}
\renewcommand{\thetable}{S\arabic{table}}

\newcounter{suppnote}
\newcommand{\suppnote}[1]{%
  \refstepcounter{suppnote}
  \subsection*{Supplementary Note \thesuppnote. #1}
}

\suppnote{Input representation example}
\label{inputwxample}
To illustrate the model input, we provide a de-identified example using the OOD prediction task. For each patient, all ICD-10 diagnosis codes from visits within the 12-month observation window were mapped to their textual descriptions and concatenated in chronological order. This diagnosis-text sequence was then provided to \textsc{OODMamba} to predict whether the patient would experience an opioid overdose event within the subsequent 180 days. The example below is illustrative only and does not correspond to a real patient, to avoid any privacy risk.

\begin{figure}[H]
\centering
\noindent\fbox{%
\begin{minipage}{0.96\linewidth}
\textbf{Task:} Predict whether the patient will experience an opioid overdose event (OOD) within the 180-day prediction window.

\textbf{Observation window:} 12 months before the first OUD diagnosis.

\textbf{Visit 1: 2021-02-14.}\\
ICD-10 codes: I10, E11.9\\
Mapped descriptions: Essential hypertension; Type 2 diabetes mellitus without complications.

\textbf{Visit 2: 2021-05-03.}\\
ICD-10 codes: M54.5, G89.29\\
Mapped descriptions: Low back pain; Other chronic pain.

\textbf{Visit 3: 2021-08-19.}\\
ICD-10 codes: F41.9, G47.00\\
Mapped descriptions: Anxiety disorder, unspecified; Insomnia, unspecified.

\textbf{Visit 4: 2021-11-27.}\\
ICD-10 codes: F11.90, Z79.891\\
Mapped descriptions: Opioid use, unspecified, uncomplicated; Long-term current use of opiate analgesic.

\textbf{Final chronological input sequence fed to the model:}\\
Essential hypertension; Type 2 diabetes mellitus without complications. Low back pain; Other chronic pain. Anxiety disorder, unspecified; Insomnia, unspecified. Opioid use, unspecified, uncomplicated; Long-term current use of opiate analgesic.

\textbf{Model output:} A binary prediction indicating opioid overdose or no opioid overdose.
\end{minipage}
}
\caption{Example de-identified diagnosis-text input for opioid overdose prediction.}
\label{fig:input_example}
\end{figure}

\suppnote{Cohort construction}

\begin{figure}[H]
    \centering
    \includegraphics[width=0.8\columnwidth]{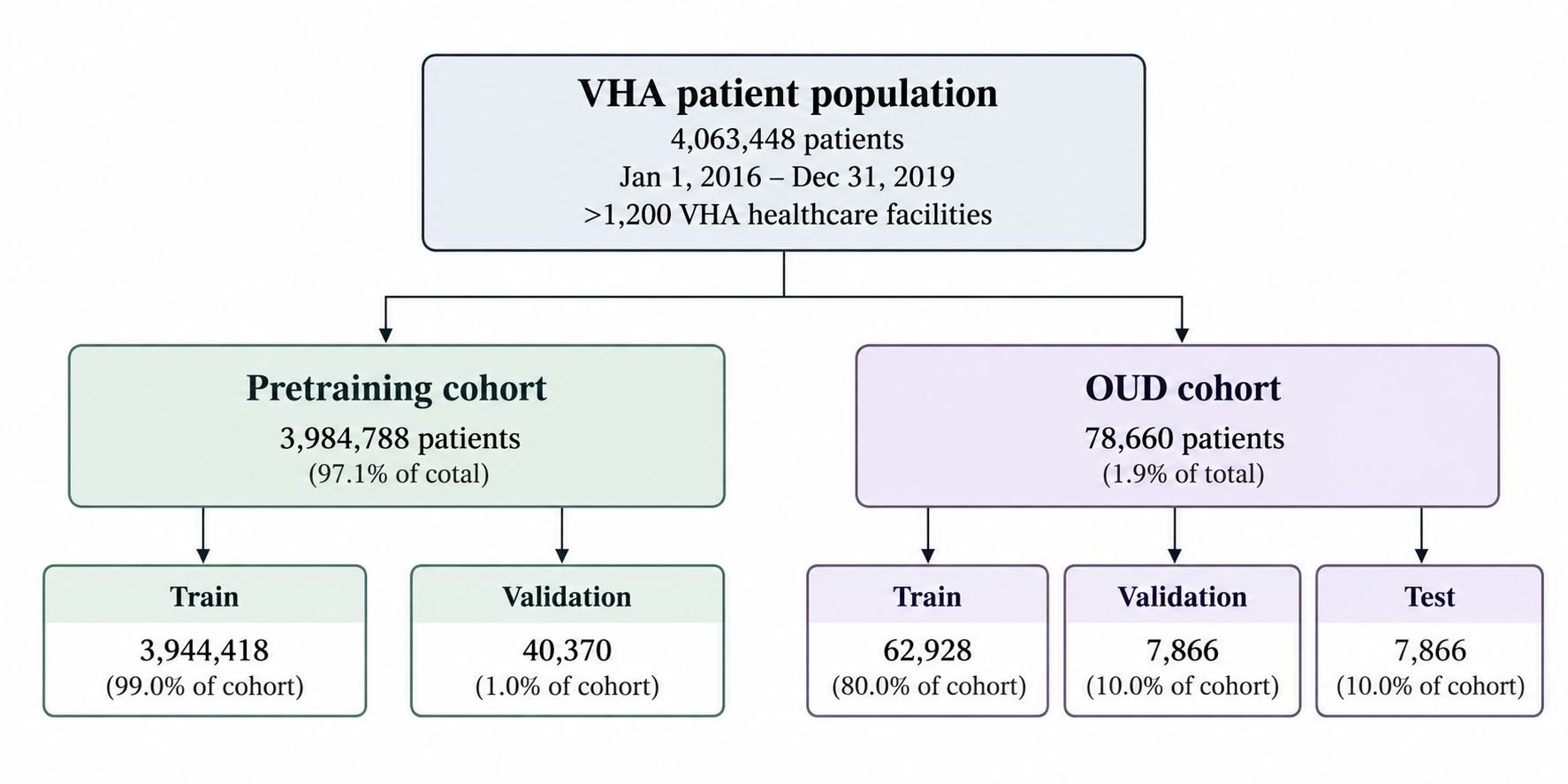}
    \caption{Cohort construction for the VHA opioid use disorder study population.}
    \label{fig:cohort}
\end{figure}

Figure~\ref{fig:cohort} illustrates the cohort construction process for the VHA opioid use disorder study population. We first identified 4,063,448 VHA patients receiving care between January 1, 2016 and December 31, 2019 across more than 1,200 VHA healthcare facilities. From this population, 3,984,788 patients were used to construct the pretraining cohort, which was split into 3,944,418 training patients and 40,370 validation patients. Separately, we identified an OUD cohort of 78,660 patients for downstream overdose prediction. This cohort was divided into training, validation, and test sets containing 62,928, 7,866, and 7,866 patients, respectively. This design separates large-scale clinical pretraining from task-specific OUD outcome modeling, while preserving held-out validation and test cohorts for unbiased evaluation.

% 
% \suppnote{Training hyperparameters}

% \begin{table}[H]
% \centering
% \small
% \begin{tabular}{ll}
% \toprule
% \textbf{Parameter} & \textbf{Setting} \\
% \midrule
% Optimizer & AdamW \\
% Learning rate & $2\times10^{-4}$ \\
% Training epochs & 20 \\
% Hardware & NVIDIA A100 (80\,GB) \\
% Batch size & 16 \\
% Warmup steps / ratio & 0 \\
% Weight decay & 0.01 \\
% Max sequence length & 2K \\
% Dropout (classification head) & 0.1 \\
% Checkpoint selection & Lowest validation loss \\
% \bottomrule
% \end{tabular}
% \caption{Training hyperparameters used for \textsc{OODMamba} and the fine-tuned language-model comparators.}
% \label{tab:hyperparameters}
% \end{table}

\suppnote{Outcome code mapping}

The primary outcome in this study was a composite EHR-ascertained opioid
overdose event within 180 days of the index OUD diagnosis. The active
single-endpoint analysis did not model fatal and non-fatal overdose separately;
therefore, the code mapping below reports the ICD-10 diagnosis-code definition
used to identify qualifying opioid-overdose events in the EHR.

\begin{table}[H]
\centering
\small
\begin{tabular}{p{4.5cm} p{8.2cm}}
\toprule
\textbf{Outcome Ascertainment} & \textbf{ICD-10 Diagnosis Codes} \\
\midrule
\textbf{EHR opioid overdose event} \newline \emph{(composite event captured within the health system)} & 
\begin{tabular}[t]{@{}l@{}}
\textbf{Heroin:} T40.1X1, T40.1X2, T40.1X3, T40.1X4 \\
\textbf{Opium:} T40.0X1, T40.0X2, T40.0X3, T40.0X4 \\
\textbf{Other opioids:} T40.2X1, T40.2X2, T40.2X3, T40.2X4 \\
\textbf{Methadone:} T40.3X1, T40.3X2, T40.3X3, T40.3X4 \\
\textbf{Other synthetic narcotics:} T40.4X1*, T40.4X2*, \\ T40.4X3*, T40.4X4*
\end{tabular} \\
\bottomrule
\end{tabular}
\caption{ICD-10 diagnosis codes used to construct the composite EHR-ascertained opioid overdose label for the active single-endpoint prediction task.}
\label{tab:icd_codes}
\end{table}

% \suppnote{Subgroup evaluation}

% \begin{table}[H]
% \centering
% \scriptsize
% \renewcommand{\arraystretch}{1.15}
% \setlength{\tabcolsep}{8pt}
% \begin{tabular}{llcccc}
% \toprule
% & & \textbf{N} & \multicolumn{3}{c}{\textbf{Opioid Overdose (OOD)}} \\
% \cmidrule(lr){4-6}
% \textbf{Group} & \textbf{Subgroup} & \textbf{(test)} & \textbf{AUPRC} & \textbf{AUROC} & \textbf{PPV} \\
% \midrule
% \multirow{7}{*}{Age}
% & 18--30 & 442     & 14.16 & 56.53 & 14.92 \\
% & 31--40 & 1{,}272 & 11.72 & 61.93 & 11.26 \\
% & 41--50 & 917     & 5.76  & 53.53 & 8.33  \\
% & 51--60 & 1{,}783 & 13.98 & 63.47 & 19.56 \\
% & 61--70 & 2{,}542 & 17.10 & 63.43 & 23.36 \\
% & 71--80 & 782     & 18.26 & 60.12 & 40.00 \\
% & 80+    & 128     & 28.35 & 60.08 & 33.33 \\
% \midrule
% \multirow{2}{*}{Sex}
% & Female & 624     & 12.60 & 61.77 & 16.13 \\
% & Male   & 7{,}242 & 13.05 & 62.37 & 16.48 \\
% \midrule
% \multirow{3}{*}{\makecell[l]{Race/\\ethnicity}}
% & Not Hispanic or Latino & 7{,}281 & 12.85 & 62.08 & 16.59 \\
% & Hispanic or Latino     & 473     & 22.37 & 66.75 & 20.68 \\
% & Other or unknown       & 112     & \multicolumn{3}{c}{\emph{Descriptive only; $<10$ positive events}} \\
% \bottomrule
% \end{tabular}
% \caption{\textsc{OODMamba} subgroup performance for opioid overdose across age, sex, and race/ethnicity strata. Metrics are reported as AUPRC, AUROC, and PPV. Strata with fewer than 10 positive events are reported descriptively only and were not used for subgroup comparisons. All metric values are reported as percentages.}
% \label{tab:demographic_subgroups}
% \end{table}